\documentclass[preprint,review,12pt]{elsarticle}

\usepackage{hyperref}

\usepackage{amsmath,amssymb,amsfonts,amsthm}

\usepackage{graphicx}

\makeatletter
\def\ps@pprintTitle{%
 \let\@oddhead\@empty
 \let\@evenhead\@empty
 \def\@oddfoot{}%
 \let\@evenfoot\@oddfoot}
\makeatother

\journal{Journal of Neural Networks}

\biboptions{authoryear}

\begin{document}

\begin{frontmatter}

\title{Towards a mathematical framework to inform Neural Network modelling via Polynomial Regression}

\author[IBiDat]{Pablo Morala\corref{mycorrespondingauthor}}
\cortext[mycorrespondingauthor]{Corresponding author at: uc3m-Santander Big Data Institute, Universidad Carlos III de Madrid. Getafe (Madrid), Spain.}
\ead{pablo.morala@uc3m.es}

\author[IBiDat]{Jenny Alexandra Cifuentes}
\author[IBiDat,Department_of_Statistics]{Rosa E. Lillo} 
\author[IBiDat]{Iñaki Ucar}

\address[IBiDat]{uc3m-Santander Big Data Institute, Universidad Carlos III de Madrid. Getafe (Madrid), Spain.}
\address[Department_of_Statistics]{Department of Statistics, Universidad Carlos III de Madrid. Getafe (Madrid), Spain.}

\begin{abstract}
    Even when neural networks are widely used in a large number of applications, they are still considered as black boxes and present some difficulties for dimensioning or evaluating their prediction error. This has led to an increasing interest in the overlapping area between neural networks and more traditional statistical methods, which can help overcome those problems. In this article, a mathematical framework relating neural networks and polynomial regression is explored by building an explicit expression for the coefficients of a polynomial regression from the weights of a given neural network, using a Taylor expansion approach. This is achieved for single hidden layer neural networks in regression problems. The validity of the proposed method depends on different factors like the distribution of the synaptic potentials or the chosen activation function. The performance of this method is empirically tested via simulation of synthetic data generated from polynomials to train neural networks with different structures and hyperparameters, showing that almost identical predictions can be obtained when certain conditions are met. Lastly, when learning from polynomial generated data, the proposed method produces polynomials that approximate correctly the data locally.
\end{abstract}


\end{frontmatter}

\section{Introduction}

Neural networks (NNs) have been established as a one of the most used models in machine learning, specially with the development of deep neural networks and their use in a wide range of applications  (\cite{lecunDeepLearning2015}. However, neural networks present several problems. Choosing the hyperparameters of NNs still depends mostly on an exploratory approach by trial and error, either for the learning algorithm parameters (\cite{bengioPracticalRecommendationsGradientBased2012}) or for their structure topology like the needed number of layers,  the number of hidden units per layer or their connections (\cite{HIROSE199161}, \cite{weymaereInitializationOptimizationMultilayer1994}, \cite{maNewTrainingStrategies2004}), with genetic algorithms as an approach that has been explored to solve this (\cite{leungTuningStructureParameters2003}).

Another of their problems is that neural networks do not directly provide an estimate of the uncertainty produced in their predictions, which is of crucial importance in most of their applications, like flood predictions (\cite{tiwariUncertaintyAssessmentEnsemble2010}), wind power forecasting (\cite{wanProbabilisticForecastingWind2014}) or molecular and atomic predictions (\cite{musilFastAccurateUncertainty2019}). In order to solve these kind of problems and quantify the uncertainty, prediction intervals have been employed, with different methods to approximate them, like bootstrap, Bayesian approaches, use of the likelihood, MVE, delta method (\cite{tibshiraniComparisonErrorEstimates1996}, \cite{khosraviComprehensiveReviewNeural2011}, \cite{kabirNeuralNetworkBasedUncertainty2018}) and with more recent approaches like LUBE (Lower Upper Bound Estimation) (\cite{khosraviLowerUpperBound2011}), where a new NN is trained to obtain the prediction intervals of the original NN. Improvements are still being made in this aspect with recent publications like (\cite{kabirOptimalUncertaintyguidedNeural2021}).

Furthermore, there are still concerns about the opaque black-box nature (\cite{benitezAreArtificialNeural1997}, \cite{shwartz-zivOpeningBlackBox2017}). There have been many contributions that intend to  solve these kind of problems, like local interpretations with a simpler model around a given prediction with LIME (\cite{ribeiroWhyShouldTrust2016a}), addressing variable importance for each prediction with SHAP (\cite{lundbergUnifiedApproachInterpreting2017a}), using influence functions from robust statistics to trace back a prediction through the learning model (\cite{kohUnderstandingBlackboxPredictions2017}) or using inversion to obtain the input that generated a given output (\cite{saadNeuralNetworkExplanation2007}). These kind of proposals fall in the framework of explainable artificial intelligence that has gained attention in recent years  (\cite{barredoarrietaExplainableExplainableArtificial2020}, \cite{guidottiSurveyMethodsExplaining2018}, \cite{adadiPeekingBlackBoxSurvey2018}), with also an important focus on neural networks (\cite{angelovExplainableDeepNeural2020}, \cite{samekInterpretableMachineLearning2020}). However, most of these approaches start with a black box model and then try to explain it, instead of using directly a simpler and interpretable model as proposed in \cite{rudinStopExplainingBlack2019}. 

In the context of solving the aforementioned problems in prediction uncertainty, interpretability or even the topological structure, there is an increasing interest in merging neural networks with more traditional statistics techniques, like Lasso regression in LassoNet (\cite{lemhadriLassoNetNeuralNetwork2020}), Neural Additive Models inspired in  generalized additive models (\cite{agarwalNeuralAdditiveModels2020}) or studying the relation between NNs and Multivariate Polynomial Regression (\cite{chengPolynomialRegressionAlternative2019}), where Polynomial Regression (PR) is postulated to be equivalent to feed forward NNs. Even when NNs are Universal Approximators (\cite{hornikMultilayerFeedforwardNetworks1989}, \cite{HORNIK1990551}, \cite{HORNIK1991251}) and PR can approximate any continuous, bounded smooth function according to the Stone-Weierstrass Theorem, the authors propose that both are indeed equivalent and that the NN learning process is creating a polynomial with higher orders from each layer. This is based on the idea that activation functions in the NN can be computed with a polynomial using Taylor series (\cite{temurtasStudyNeuralNetworks2004}). While this equivalence is not explicitly proven in (\cite{chengPolynomialRegressionAlternative2019}), the experimental results conclude that in certain situations, PR performs as accurately or even better than NNs, indicating that these idea could be explored to build some mathematical tools and intuitions to help solving some of the NN problems. There are also other examples of studies where NNs and PR are explored together. In \cite{liuPolynomialNeuralNetwork}, Taylor expansion is computed through a neural network, learning the coefficients of the Taylor expansion through backpropagation. Alternatively, \cite{liuDeepNeuralNetworkBased2019} presents a deep neural network to solve iterative algorithms by means of the composition of blocks of single hidden layer NNs using PR as a bridge between these blocks.  

In this paper, the proposal of \cite{chengPolynomialRegressionAlternative2019} is deeply analyzed to try to build a mathematical framework that could be useful to inform and obtain a better understanding of NNs and provide new ways to solve the problems of uncertainty,  interpretability or structure tuning. An explicit formula is built, which takes the weights of a feed-forward NN with a single hidden layer and a single output and obtains the coefficients of a PR that approximates the NN. This mathematical relation between the NN and the obtained PR could help transferring knowledge from PR to NNs, in order to solve the tuning of parameters like the number of hidden units or the number of layers, the determination of prediction intervals to asses the uncertainty and also allowing for a new  framework that helps interpreting NNs. Furthermore, the equivalence implies that in applications where a NN of the given characteristics performs accurately, a PR model could be trained in first place in a faster way with less hyperparameter tuning.

This paper is structured as follows. In Section \ref{proposed_method} the proposed method is presented, introducing first the notation that will be used, building then the formula used to obtain the polynomial coefficients of the approximation and finally discussing the validity of the Taylor approximation in some of the most common activation functions. In Section \ref{experimental_results}, the experimental results obtained by simulations are presented and discussed. Finally, in Section \ref{conclusion}, conclusions of this work are outlined and future works that could emerge from it are consequently discussed.

\section{Proposed method}\label{proposed_method}
\subsection{Notation}

Consider a feed-forward neural network with a single hidden layer with $h_1$ units and a single output. The activation function at the output will be linear, and the activation function in the hidden layer will be denoted as $g()$. Denote the input to the neural network as $\mathbf{x}=(x_0,x_{1}, \dots, x_{p})$, where $x_0=1$ is a constant term. The weights connecting the input with the hidden units will be $\mathbf{w_j}=(w_{0,j},w_{1,j}, \dots, w_{i,p})$. At each hidden unit $y_j$ the output is computed as:
\begin{equation}\label{y_neuron}
    y_j=g\left(u_j \right)=g\left(\sum_{i=0}^p w_{i,j} x_{i} \right),
\end{equation}
where the synaptic potentials are $u_j=\sum_{i=0}^p w_{i,j} x_{i}$. The final output $z$ of the NN is computed as a linear combination of the hidden layers:
\begin{equation}\label{z_output}
    z=\sum_{j=0}^{h_1} v_{j} y_{j} = \sum_{j=0}^{h_1} v_{j} g\left(\sum_{i=0}^p w_{i,j} x_{i} \right).
\end{equation}
The notation used for a polynomial regression with response $Y$ and its coefficients is:
\begin{multline}\label{general_regression} Y=\beta_{0} + \beta_{1} x_{1} + \dots + \beta_{p}x_{p} + \dots + \\ \beta_{11} x_{1}^2 + \beta_{12} x_{1}x_{2} +\dots + \beta_{1 p\dots p}x_1x_p^{k-1} + \beta_{p\dots p}x_p^{k}.
\end{multline}

The problem addressed in this work is finding the coefficients of a polynomial regression that performs equivalently as a given trained neural network. Taylor expansion allows us to approximate the activation functions of the neural network $g()$, so
\begin{equation*}
y_j=g\left(\sum_{i=0}^p(w_{i,j}x_i)\right)=\sum_{n=0}^{\infty}\frac{g^{(n)}(a)}{n!}\left(\sum_{i=0}^p(w_{i,j}x_i)-a\right)^n,
\end{equation*} for all $j=1,2,\dots,h_1$. Using the binomial theorem, the following term can be expanded:
\begin{equation*}
    \left(\sum_{i=0}^p w_{i,j}x_{i}-a\right)^n = \sum_{k=0}^n \binom{n}{k}(-a)^{n-k}\left(\sum_{i=0}^p(w_{i,j}x_i)\right)^k.
\end{equation*}
And then using the multinomial theorem:
\begin{equation*}
    \left(\sum_{i=0}^p w_{i,j}x_{i}\right)^k=\sum_{m_{0}+\cdots+m_{p}=k}{\binom{k}{m_{0}, \dots, m_{p}}} (w_{0,j}x_{0})^{m_{0}} \cdots (w_{p,j}x_{p})^{m_{p}},
\end{equation*}
where the multinomial coefficient is defined as:
\begin{equation*}
    {\binom{k}{m_{0},  \dots, m_{p}}}= \frac{k !}{m_{0}! \cdots m_{p} !},
\end{equation*}
and where the coefficients $m_i$ for $i=0,\dots,p$ represent the degree of the term $w_{i,j}x_i$, keeping in mind that they have to satisfy that $m_{0}+\cdots+m_{p}=k$. Finally, in order to have a practical implementation of this method, the infinite Taylor expansion can be truncated at a given degree $q$. Then, the general Taylor expansion of an activation function to compute the output of a hidden neuron unit is:
\begin{multline}\label{y_j_general}
    y_j=\sum_{n=0}^{q}\frac{g^{(n)}(a)}{n!}\sum_{k=0}^n \binom{n}{k}(-a)^{n-k}\times\\\times\left[\sum_{m_{0}+\cdots+m_{p}=k}{\binom{k}{m_{0}, \dots, m_{p}}} (w_{0,j}x_{0})^{m_{0}} \cdots (w_{p,j}x_{p})^{m_{p}}\right].
\end{multline}

With this expression, the output of the neural network can be obtained by taking the linear combination $z=\sum_{j=0}^{h_1} v_{j} y_{j}$. However, in the chosen approach, a Taylor expansion centered at $0$ has been selected because it simplifies the computations and also the synaptic potentials $u_j$ can be expected to be symmetrical around zero depending on the scaling method that is applied to the input data. Then, the output of a hidden unit choosing $a=0$ is:
\begin{equation*}
    y_j=\sum_{n=0}^{q}\dfrac{g^{(n)}(0)}{n!}\left[\sum_{m_{0}+\cdots+m_{p}=n}{\binom{n}{m_{0}, \dots, m_{p}}} (w_{0,j}x_{0})^{m_{0}} \cdots (w_{p,j}x_{p})^{m_{p}}\right],
\end{equation*}
and including this in the final output of the neuron from Eq.~\eqref{z_output} is:

\begin{multline*}
    z=v_0+\sum_{j=1}^{h_1}v_j\sum_{n=0}^{q}\dfrac{g^{(n)}(0)}{n!}\times\\\times\left[\sum_{m_{0}+\cdots+m_{p}=n}{\binom{n}{m_{0}, \dots, m_{p}}} (w_{0,j}x_{0})^{m_{0}} \cdots (w_{p,j}x_{p})^{m_{p}}\right].
\end{multline*}

With this final expression, this model can be associated to a polynomial regression  model like the one in Eq.~\eqref{general_regression} and consequently, the explicit expression of the $\beta$ coefficients of the regression can be obtained in terms of the weights of the neural network, keeping in mind that it is an approximation and each $\beta$ coefficient can contribute to an error. Then, considering the general polynomial model,  the intercept is denoted by:
\begin{equation}\label{betas_intercept}
    \beta_0=v_0+\sum_{j=1}^{h_1}v_j\left(\sum_{n=0}^{q}\frac{g^{(n)}(0)}{n!}(w_{0,j})^{n}\right),
\end{equation}
and the rest of the coefficients associated to any combination of variables of order $t$ are:
\begin{equation}\label{betas_formula}
    \beta_{l_1l_2\dots l_t}=\sum_{j=1}^{h_1}v_j\left(\sum_{n=t}^{q}\frac{g^{(n)}(0)}{(n-t)!\cdot m_1!\cdots m_p!} (w_{0,j})^{n-t}(w_{1,j})^{m_1}\dots (w_{p,j})^{m_p}\right),
\end{equation}
where $m_j=\sum_{i=1}^t \delta_{l_i,j}$, that is, the number of times that the coefficient $j$ appears in the $t$ indexes of $\beta_{l_1l_2\dots l_t}$. Note that the $t$ parameter appears because, in the expansion of terms of order lower than $t$, there can not appear combinations of variables of order $t$.

\subsection{Validation assessment of the Taylor expansion method}\label{Taylor_validation_section}

Before showing the results obtained when using this method, the behavior of the most commonly used activation functions are discussed in this section. The formula derived previously will work properly when the value of $\sum_{j=0}^p w_{ij}x_j$ is contained within an acceptable range around 0. This leads to the question of what is an acceptable range for a given activation function~$g$.

Here, three continuous activation functions are analyzed: the hyperbolic tangent (tanh), the sigmoid and the softplus, that are among the most used activation functions in NNs implementations (\cite{balajivenkateswaranNeuralNetworks2017}.

\begin{itemize}[]
\item{\textit{Softplus:}} The softplus function, also called \textit{smoothReLU}, is defined as a smoother version of the ReLU function, and therefore it is differentiable. Its expression is defined as: $g(x)=\ln\left(1+e^x\right)$, and its Taylor expansion up to order 8 is:
\begin{equation*}
    g(x)\approx \ln(2) +\frac{x}{2} +\frac{x^2}{8} -\frac{ x^{4}}{192} +\frac{ x^{6}}{2880} -\frac{17 x^{8}}{645120} +O\left(x^{10}\right),
    \end{equation*}
where the odd terms are 0, except for the one of order 1. In  Fig.~\ref{taylor_softplus}, the approximation and its corresponding error behaviour are shown for even values $q$.

\begin{figure}[ht]
	\centering
    \includegraphics[width = \textwidth]{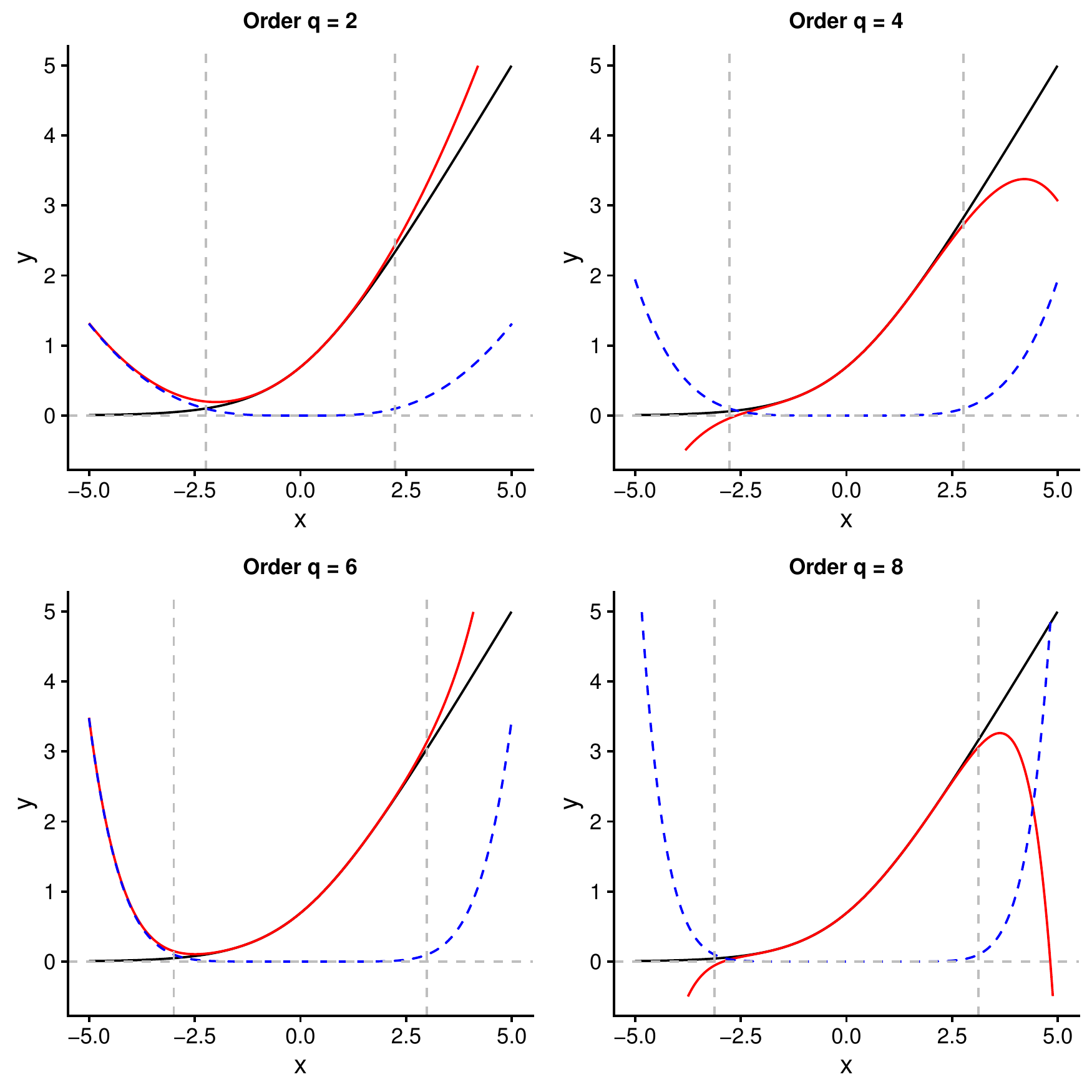}
	\caption{Softplus Taylor approximation around 0. The black line represents the softplus function, the red one the Taylor approximation for the given order $q$ and the dashed blue one the error between both functions at each point, computed as the absolute value of their difference. The vertical dashed grey lines represent the points at which the error is higher than $0.1$, as a representative value.}\label{taylor_softplus}
\end{figure}

\item{\textit{Hyperbolic tangent:}} The hyperbolic tangent is also one of the most used activation functions and it behaves as an smooth version of the step function. It restricts the output of the neuron between $-1$ and $1$ due to its expression: $g(x)=\tanh(x)=\dfrac{e^x-e^{-x}}{e^x+e^{-x}}$, and its Taylor expansion up to order $7$ is:
\begin{equation*}
    g(x)\approx x-\frac{x^{3}}{3}+\frac{2 x^{5}}{15}-\frac{17 x^{7}}{315}+O\left(x^{9}\right),
\end{equation*} where it can be seen that the even terms of the expansion are zero. Fig.~\ref{taylor_tanh} shows how the approximation and its error behaves for several odd values of $q$.

\begin{figure}[ht]
	\centering
	\includegraphics[width = \textwidth]{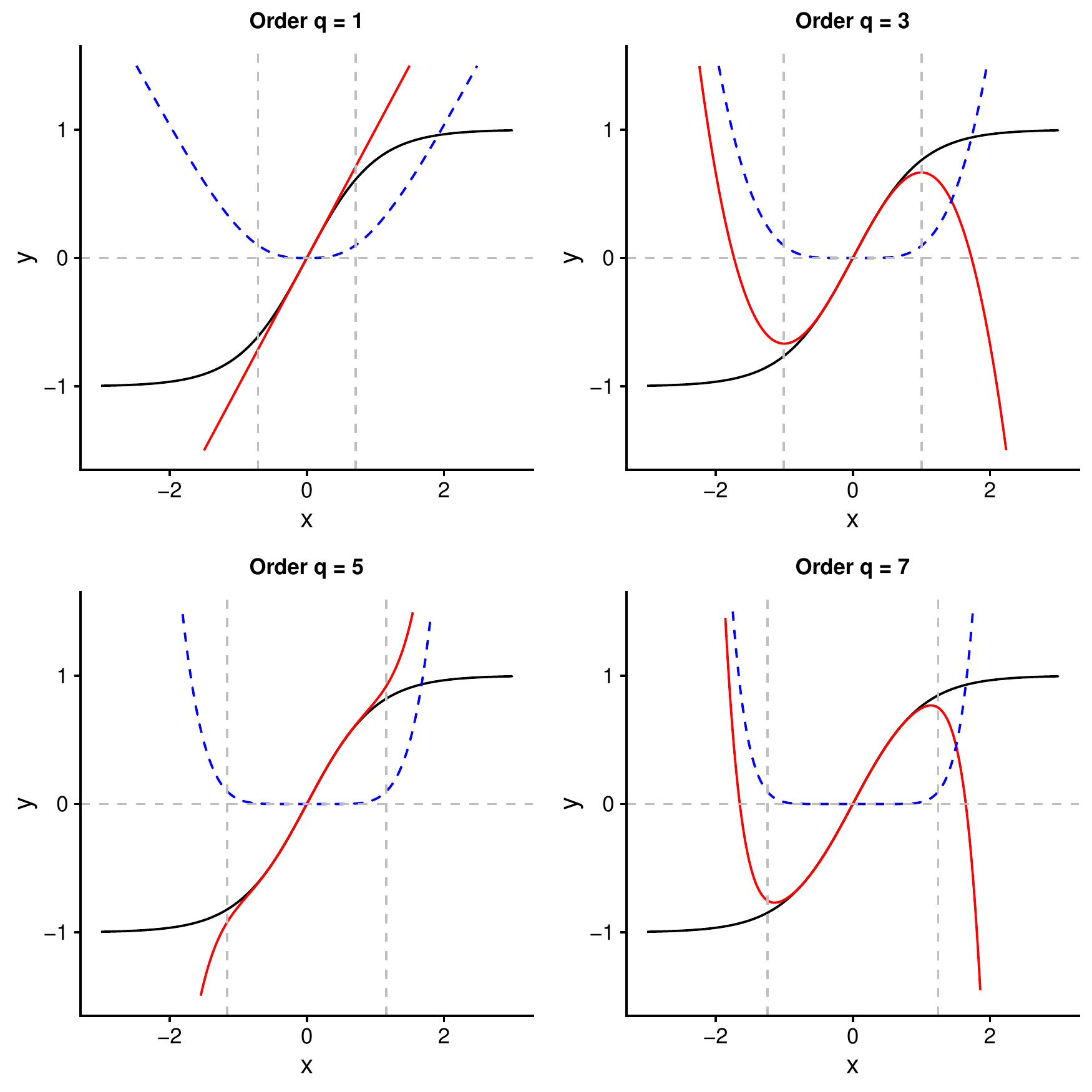}
	\caption{Hyperbolic tangent Taylor approximation around 0. The black line represents the $\tanh$ function, the red one the Taylor approximation for the given order $q$ and the dashed blue one the error between both functions at each point, computed as the absolute value of their difference. The vertical dashed grey lines represent the points at which the error is higher than $0.1$, as a representative value.}\label{taylor_tanh}
\end{figure}

\item{\textit{Sigmoid:}} The sigmoid function is similar to the hyperbolic tangent because it limits the output between  $0$ and $1$, but it has a less steep slope. It is defined as $g(x)=\dfrac{1}{1+e^{-x}}$, and its Taylor approximation up to order $7$ is:
\begin{equation*}
    g(x)\approx \frac{1}{2} + \frac{x}{4} -\frac{x^{3}}{48}+\frac{ x^{5}}{480}-\frac{17 x^{7}}{80640}+O\left(x^{9}\right),
\end{equation*} where it can be seen that the odd terms are zero, except for the constant term. Its approximation and error behaviour for several odd values of $q$ is shown in Fig.~\ref{taylor_sigmoid}.

\begin{figure}[ht]
	\centering
	\includegraphics[width = \textwidth]{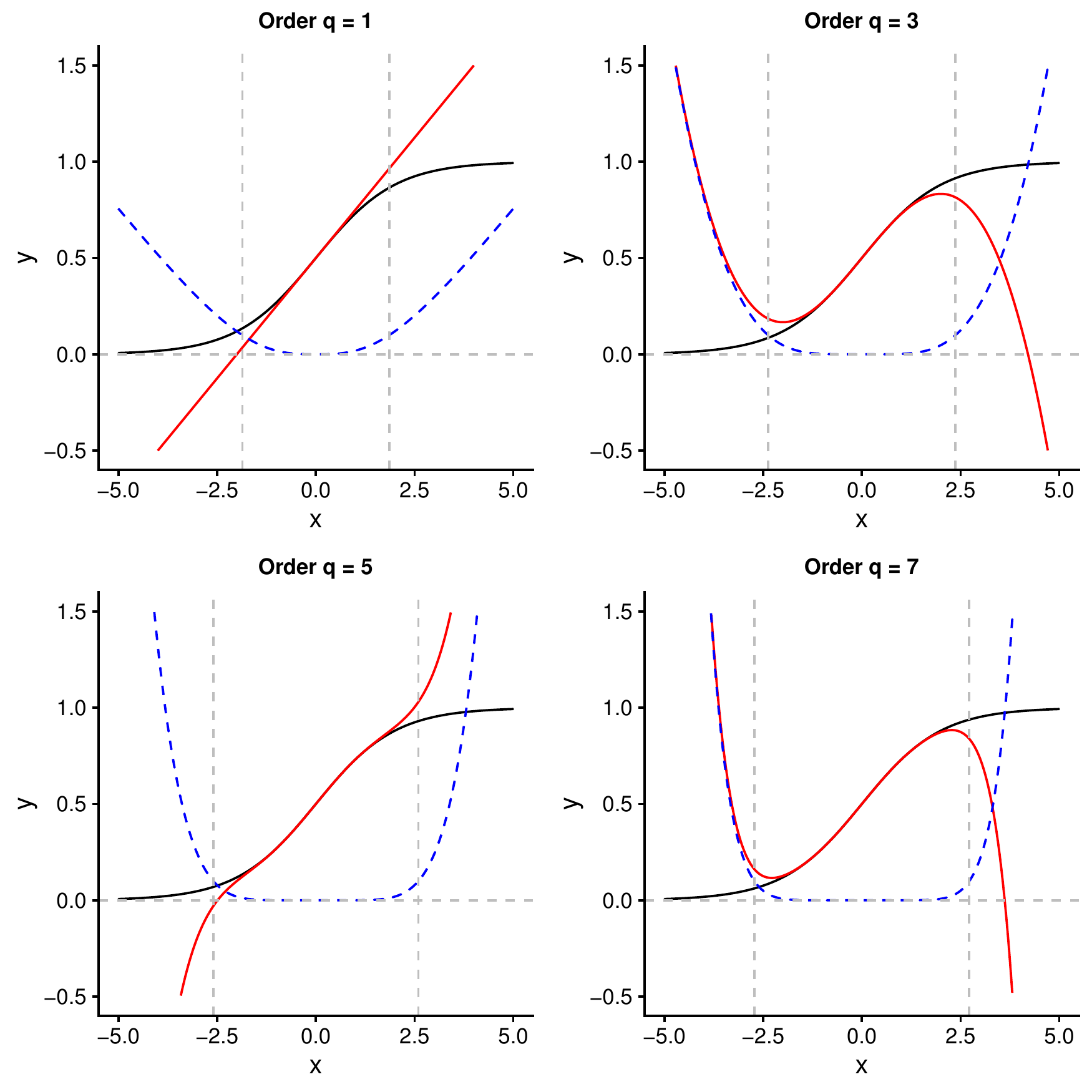}
	\caption[Sigmoid Taylor approximation around 0.]{Sigmoid Taylor approximation around 0. The black line represents the sigmoid function, the red one the Taylor approximation for the given order $q$ and the dashed blue one the error between both functions at each point, computed as the absolute value of their difference. The vertical dashed grey lines represent the points at which the error is higher than $0.1$, as a representative value.}\label{taylor_sigmoid}
\end{figure}

\end{itemize}

\section{Simulation study and discussion}\label{experimental_results}

The main experimental results obtained through different simulations are presented in this section, showing the performance of the obtained polynomial regression with the coefficients formula from Eq.~\eqref{betas_formula}. All the code needed to perform these simulations and produce these figures is available on GitHub\footnote{https://github.com/moralapablo/nntopr}. The \textbf{general framework} used  when generating examples, unless stated otherwise, is the following, represented also in Fig.~\ref{experimental_framework_diagram}:

\begin{figure}[ht]
    	\centering
    	\includegraphics[width = \textwidth]{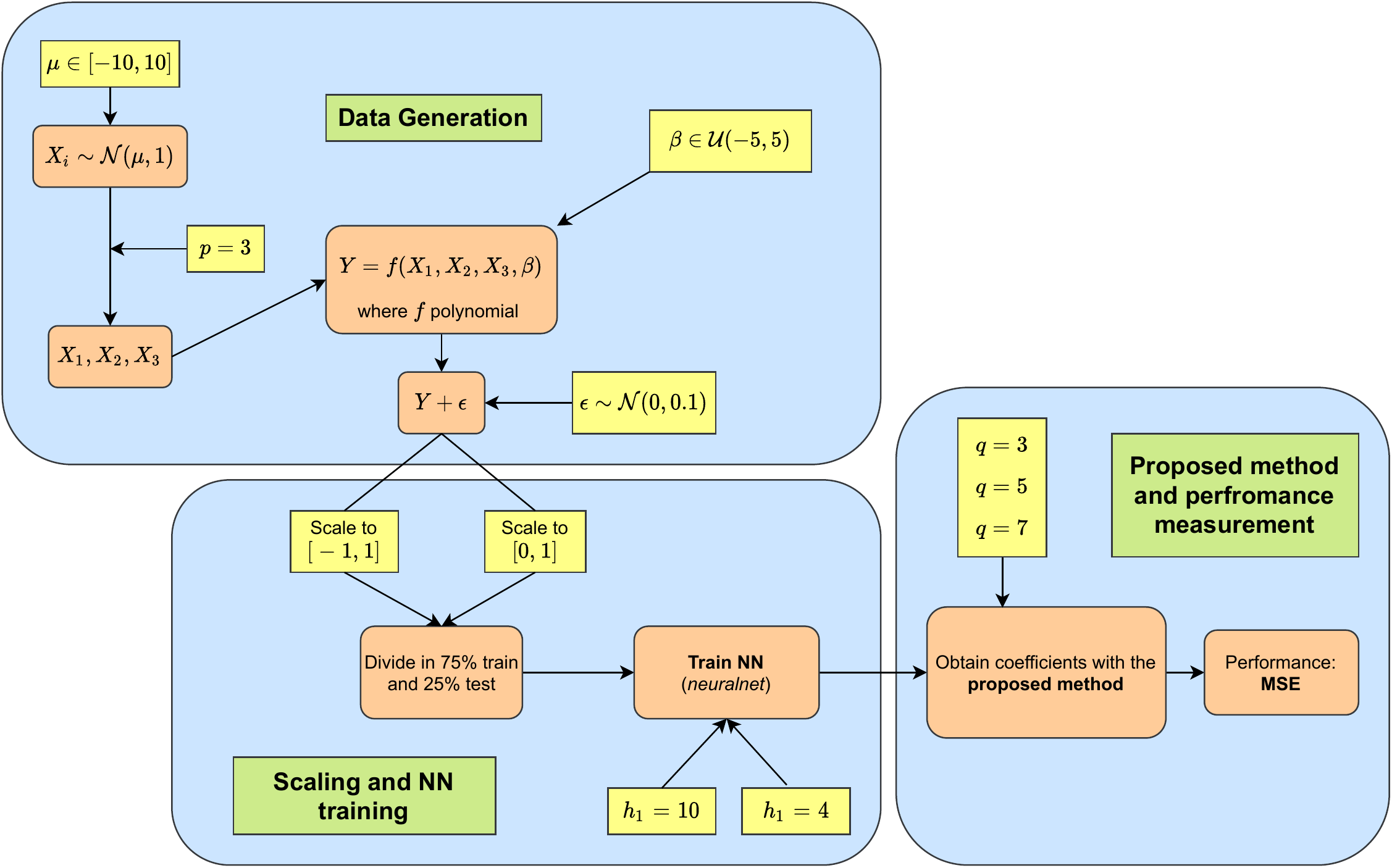}
    	\caption{Diagram representing the steps take in the general framework of the experimental simulations}
    	\label{experimental_framework_diagram}
    \end{figure}

\begin{itemize}
    \item \textit{Data generation}: 200 samples of $p=3$ independent normal distributions $X_i$ for $i=1,2,3$ were generated, with random mean values, for each component obtained, from an uniform distribution between $-10$ and $10$ and variance equal to $1$. Then, for a given degree $q_{data}=2$ the response $Y$ was computed as a polynomial regression of that order. To do this, the needed number of $\beta$ coefficients were generated for each of the terms of the polynomial regression from an uniform distribution between $-5$ and $5$. Finally, a normal error to the response variable $Y$ with mean $ \mu = 0$ and standard deviation $\sigma = 0.1$ was included.
    
    Note that these intervals for the values of  $X_i$ and the coefficients $\beta$, the dimension $p$ and the original degree $q$ are chosen arbitrarily but represent simple regression cases that can be used as a benchmark for our formula. Recall that Eq.~\eqref{betas_formula} does not depend directly on the data used (aside from the dimension $p$), but rather will approximate the response of the NN independently of how well the NN represents the data.
     
    \item \textit{Scaling the data and training the NN}: In order to train the NN, the data were scaled first by choosing from two different possibilities, scaling the data in the interval $[0,1]$ and scaling the data in the interval $[-1,1]$.  The results will be discussed later for each method.
    
    After the scaling is done, the data was split into train (75\%) and test (25\%) sets. Then, the NN was trained with the train data, with a resilient backpropagation with weight backtracking algorithm (\cite{fritschNeuralnetTrainingNeural2019}, \cite{RPROP_algorithm}), choosing the desired activation function (softplus, tanh or sigmoid) and the number of neurons $h_1$ in the single hidden layer (10 and 4 were the chosen values). Also the output of the neural network is chosen to be linear to solve the regression problem and to satisfy the hypothesis of our calculations.
    
    \item \textit{Building the PR:} With the NN trained, the weights $w_{i,j}$ and $v_j$ were extracted from the NN output and, after fixing a certain value for $q$ (3, 5 or 7), the $\beta$ coefficients were computed up to the chosen degree $q$ through the implementation of  Eq.~\eqref{betas_formula}.
    
    \item \textit{Performance measurement:} Based on the coefficients computation, the performance of the method is evaluated by comparing the predictions of the NN with the predictions of the obtained PR. In the simulation this is mainly done with the mean squared error (MSE) between these two predictions.
    
    Visual representations were used as well to compare these two predictions. If the PR is truly representing the NN, the plot will show an straight line close to $y=x$, plotted in red. To have an idea of how well the NN predicts the original response $Y$ of the test data, it is also plotted. Recall again that the important part here is to see if the PR accurately represents the NN independently of how well the NN performs with respect to the original data.
\end{itemize}

\subsection{Performance examples of the proposed method}

\begin{figure}[ht]
    	\centering
    	\includegraphics[width = \textwidth]{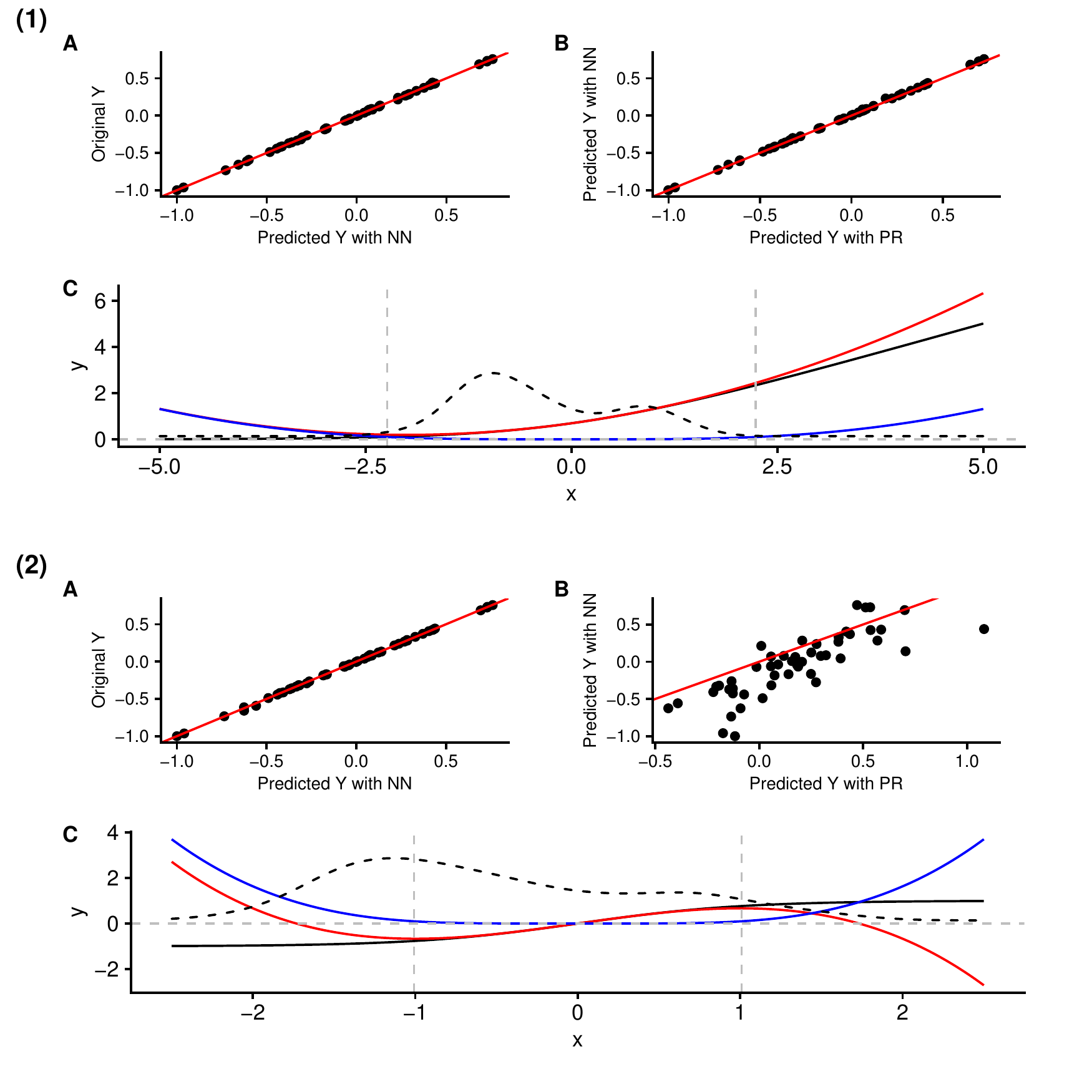}
    	\caption{Performance examples of the proposed method. \textbf{(1)}: Example with softplus activation function, $h_1=4$, $q=3$, data scaled to the interval $[-1,1]$. \textbf{(2)}: Example with hyperbolic tangent activation function, $h_1=4$, $q=3$, data scaled to the interval $[-1,1]$. In both examples: \textbf{(a)} Performance of the NN predictions with respect to the original response Y in the test data set. The red line shows $y=x$, where the dots should land in a perfect scenario. \textbf{(b)} Same plot as the previous one but comparing the predictions of the associated PR with the ones of the NN. \textbf{(c)} Taylor approximation (red) to the original function (black), its error (blue) and the distribution of the input values (dashed)}
    	\label{performance_examples}
    \end{figure}

Two typical examples of how the proposed method works and its limitations are presented now. In these examples (Fig.~\ref{performance_examples}) the following items are represented: the neural network performance with respect to the original response (\textbf{A}), the performance of the polynomial regression predictions with respect to the predictions made with the neural network (\textbf{B}) and then a plot (\textbf{C}) showing the real activation function (black), the Taylor approximation with the given degree $q$ (red) and the distribution of the synaptic potentials $u_j=\sum_{i=0}^p w_{i,j}x_i$, that the activation function receives for each of the $j = 1,\dots, h_1$ neurons in the hidden layer, computed for each of the samples in the test data set. This scheme is presented for two different examples. Example 1 corresponds to a case in which the proposed method works properly, obtaining a PR that is almost equivalent in the predictions to the NN. This appropriate behaviour is obtained due to the fact that the input values fall in a range in which the Taylor expansion behaves properly. On the other hand, example 2 corresponds to a case where the obtained PR is not representing correctly the predictions of the NN and therefore the proposed method is failing. However, this is explained by the fact that a large part of the synaptic potentials distribution is falling outside the region around zero where the Taylor approximation is correct and the Taylor series diverges from the actual function values. 
 
With these two examples it can be seen that the proposed method can work but it is needed, as expected, that the inputs to the hidden neurons (dependent on the weights and the input variables of the model) are inside the acceptable approximation range of the Taylor expansion. Therefore, the performance has a random component, even in examples that had the exact same data sets fixed, due to how the NN is trained and the obtained weights. To overcome this problem, in Section \ref{MSE_subsection}, several simulations are presented to study the distributions of their performance instead of studying single examples.

\subsection{MSE between PR and NN}\label{MSE_subsection}

\begin{figure}[ht]
	\centering
	\includegraphics[width = \textwidth]{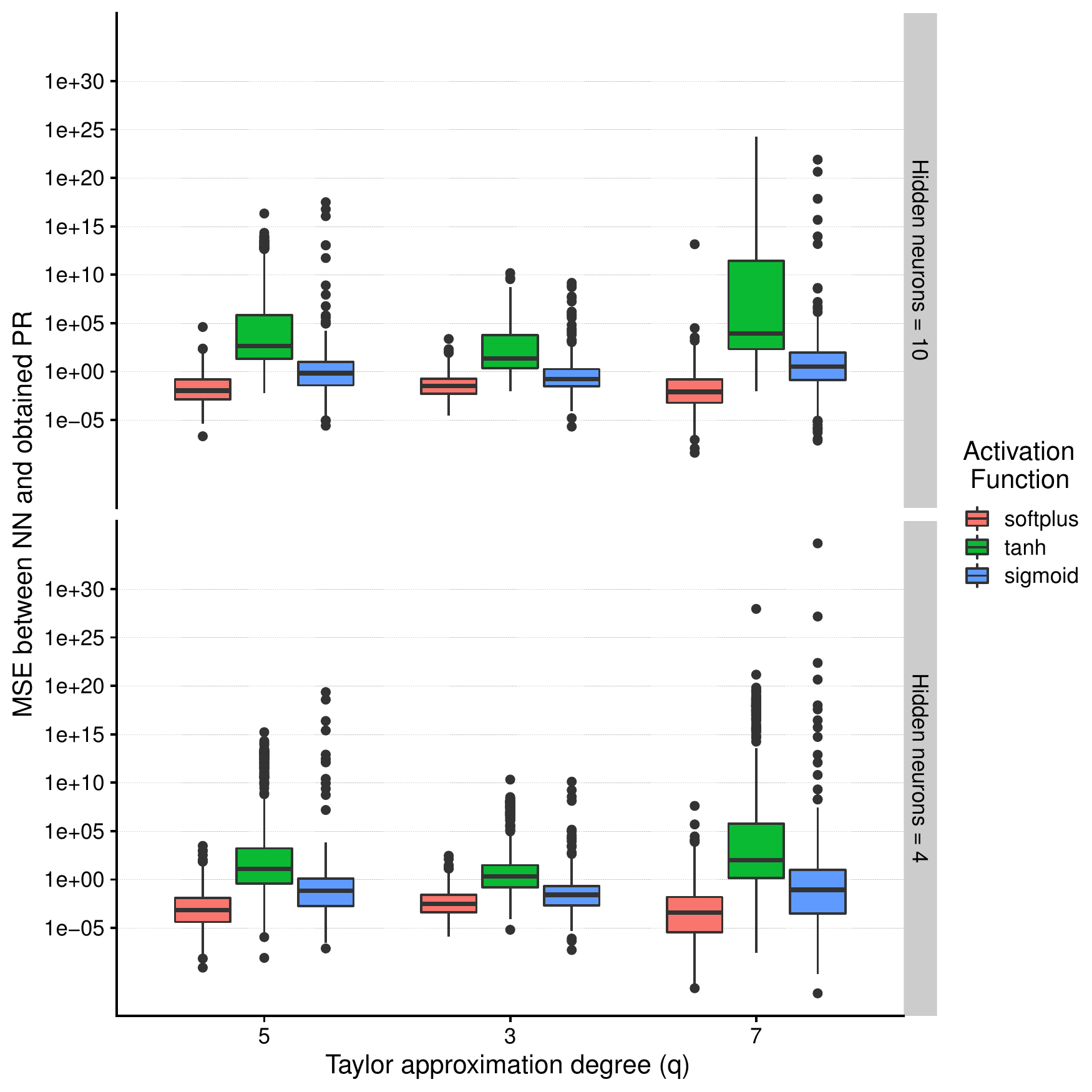}
	\caption[MSE distributions for simulations scaling in the interval \[0,1\].]{MSE distributions for simulations of 500 repetitions, scaling in $[0,1]$, changing $q$, $h_1$ and the activation function.}
	\label{boxplots1}
\end{figure}

\begin{figure}[ht]
	\centering
	\includegraphics[width = \textwidth]{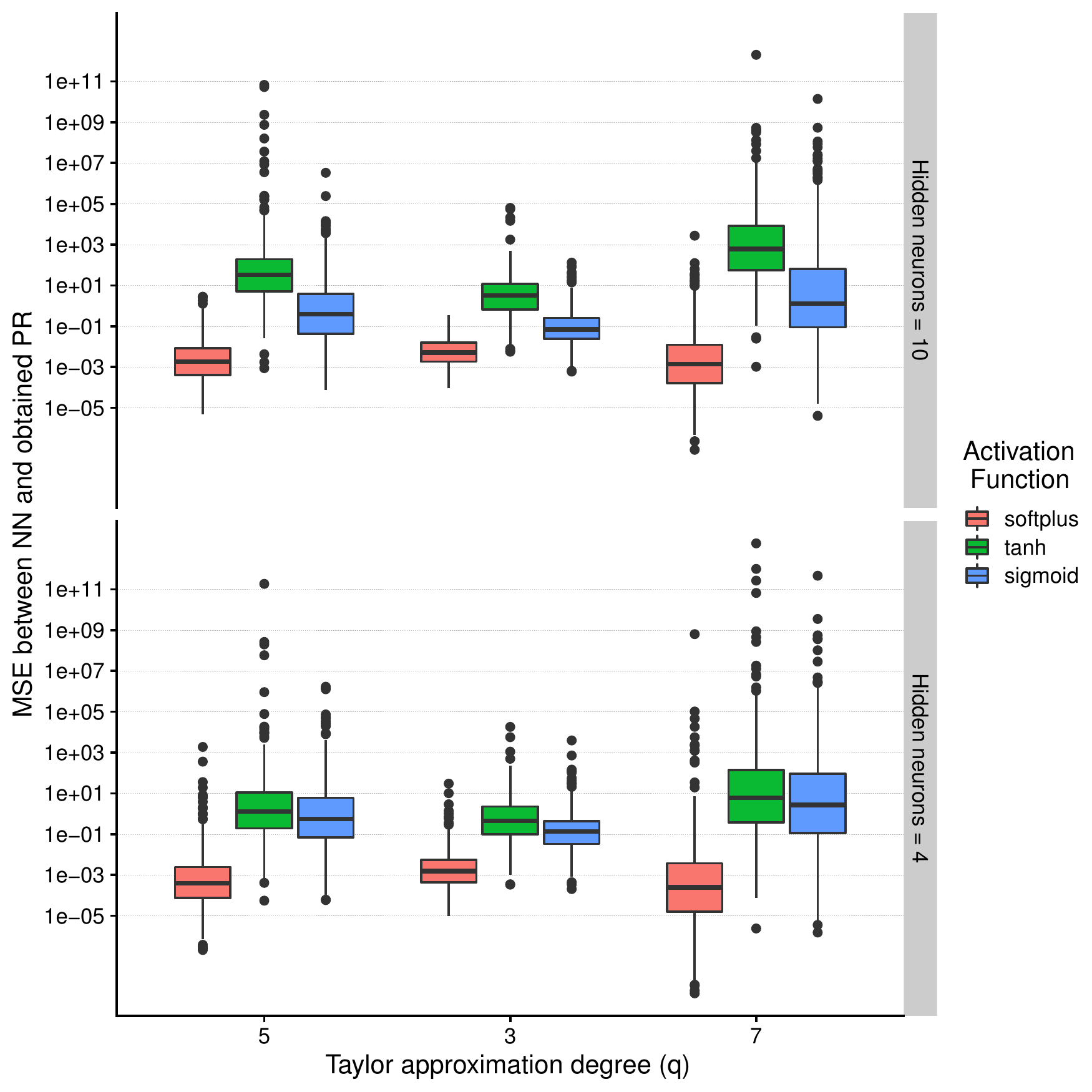}
	\caption[MSE distributions for simulations scaling in the interval \[-1,1\].]{MSE distributions for simulations of 500 repetitions, scaling in $[-1,1]$, changing $q$, $h_1$ and the activation function.}
	\label{boxplots2}
\end{figure}

Here it is presented the distribution of the mean squared error (MSE) between the values predicted by the NN and the values predicted by the associated PR obtained with our formula when simulating 500 different examples for different combinations of parameters, where each of the simulations is performed under the general framework.
Two different results are presented, Fig.~\ref{boxplots1} where the data is scaled to $[0,1]$, and Fig.~\ref{boxplots2} where it is scaled to the interval $[-1,1]$. In all of them, the examples for each activation function were trained, for $q=3,5,7$ and for $h_1=4,10$.

Comparing both Figures, the first thing that it is possible to observe is that the softplus function outperforms always the other activation functions, with the hyperbolic tangent being the worst one. The sigmoid one however shows in some cases the largest variance in its distribution. This is explained by how the Taylor expansion method behaves for these activation functions as seen in Section \ref{Taylor_validation_section}. In the hyperbolic tangent case, with a fixed maximum error, the range that does not exceed it is smaller than in the softplus or the sigmoid cases. Between the sigmoid and the softplus, the latter one has a slightly larger acceptable range and the error does not grow asymptotically so fast as in the sigmoid case.

Regarding the number of neurons, there are slightly better results when using the lower value $h_1=4$ instead of $h_1=10$. This can be explained due to the fact that with a lower number of hidden neurons, the number of degrees of freedom of the model is reduced and, in general, the algorithm training the NN will lead to smaller weights.

In terms of the Taylor degree $q$, the mean of the MSE distribution seems to be a bit lower with higher values for $q$, but it highly increases the variability of the distribution for higher values of $q$, mainly due to the more asymptotic behaviour of the error when  $q$ is increased in the Taylor expansion.

Finally, regarding the scaling method, the best method seems to be scaling to the $[-1,1]$ interval. This might be because the input values that the activation function receives are closer to 0 when the input variables of the NN are centered around 0.

\subsection{Comparison with the original polynomial}

Finally, in this section, some insights on how well the obtained polynomial represents the original polynomial that generated the data are presented. This is an important observation to make when trying to assess the interpretabilty of the model. To address the fact that the obtained coefficients of the polynomial are dependent on the randomness involved in the NN training,  four examples are generated with the same data and same hyperparameters but using a different random seed. The \textit{polyreg} package is also used to compare these results to a true polynomial regression solution.

The data is generated changing the dimensions of our general framework from $p=3$ to $p=2$, in order to make visualization possible in three dimensions. The chosen activation function is the softplus, the scaling is made in the interval $[-1,1]$, the number of hidden layers is $h_1=4$ and the order used in the Taylor approximation is $q=2$, while the data was originated from a polynomial of order $2$.

The coefficients obtained with the proposed method are computed in the space scaled to the the interval $[-1,1]$ because the weights of the NN are generated using that scaled date. Therefore, these coefficients need to be scaled back to compare them with the original polynomial that originates the data.

\begin{figure}[ht]
	\centering
	\includegraphics[width = \textwidth]{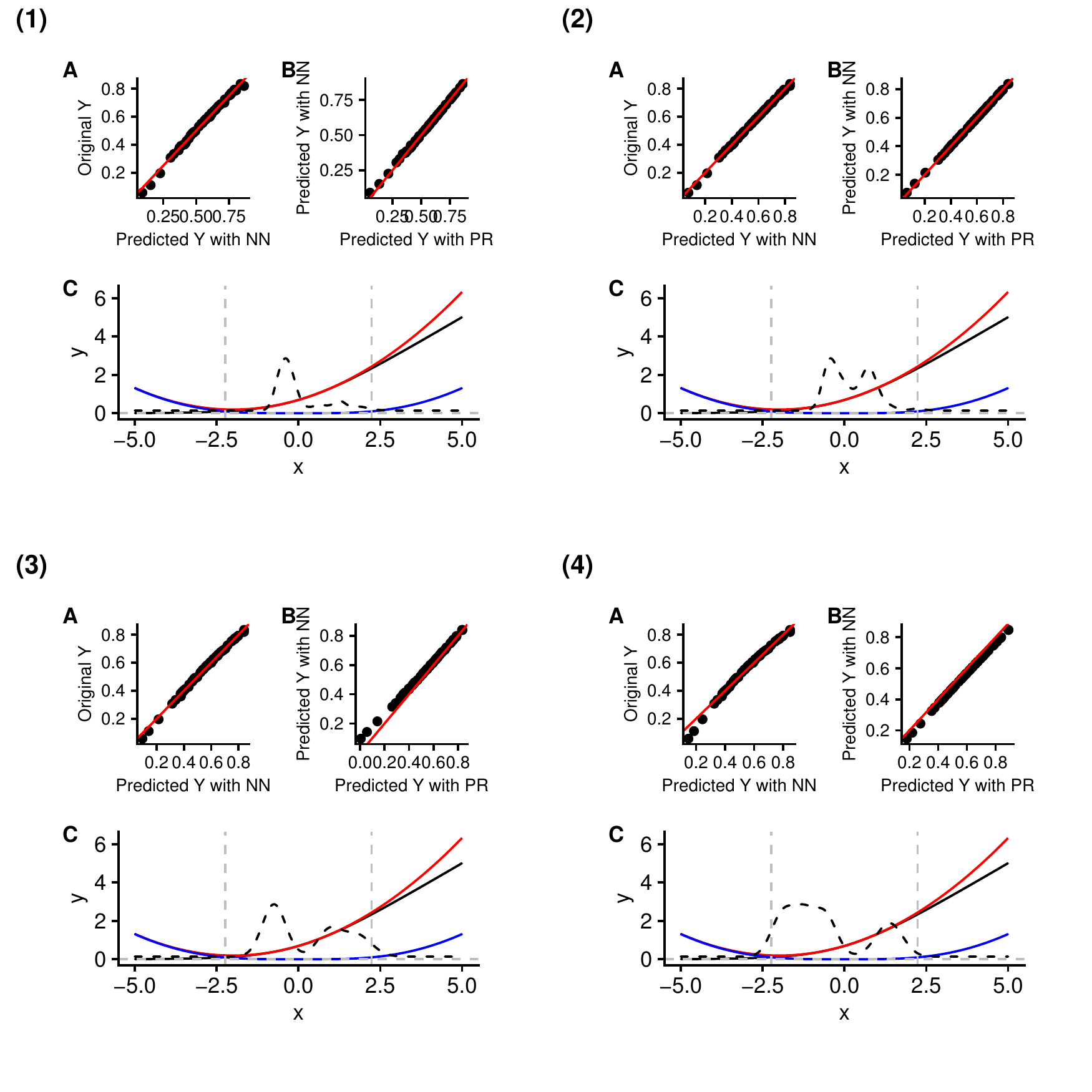}
	\caption{Four performance examples of the proposed method using the same original data but different NN initialization. Data generated from a second order polynomial. NN with $h_1=4$, $q=2$, softplus activation function and data scaled to $[-1,1]$.}
	\label{coef_comparison_performance}
\end{figure}

\begin{figure}[ht]
	\centering
	\includegraphics[width = \textwidth]{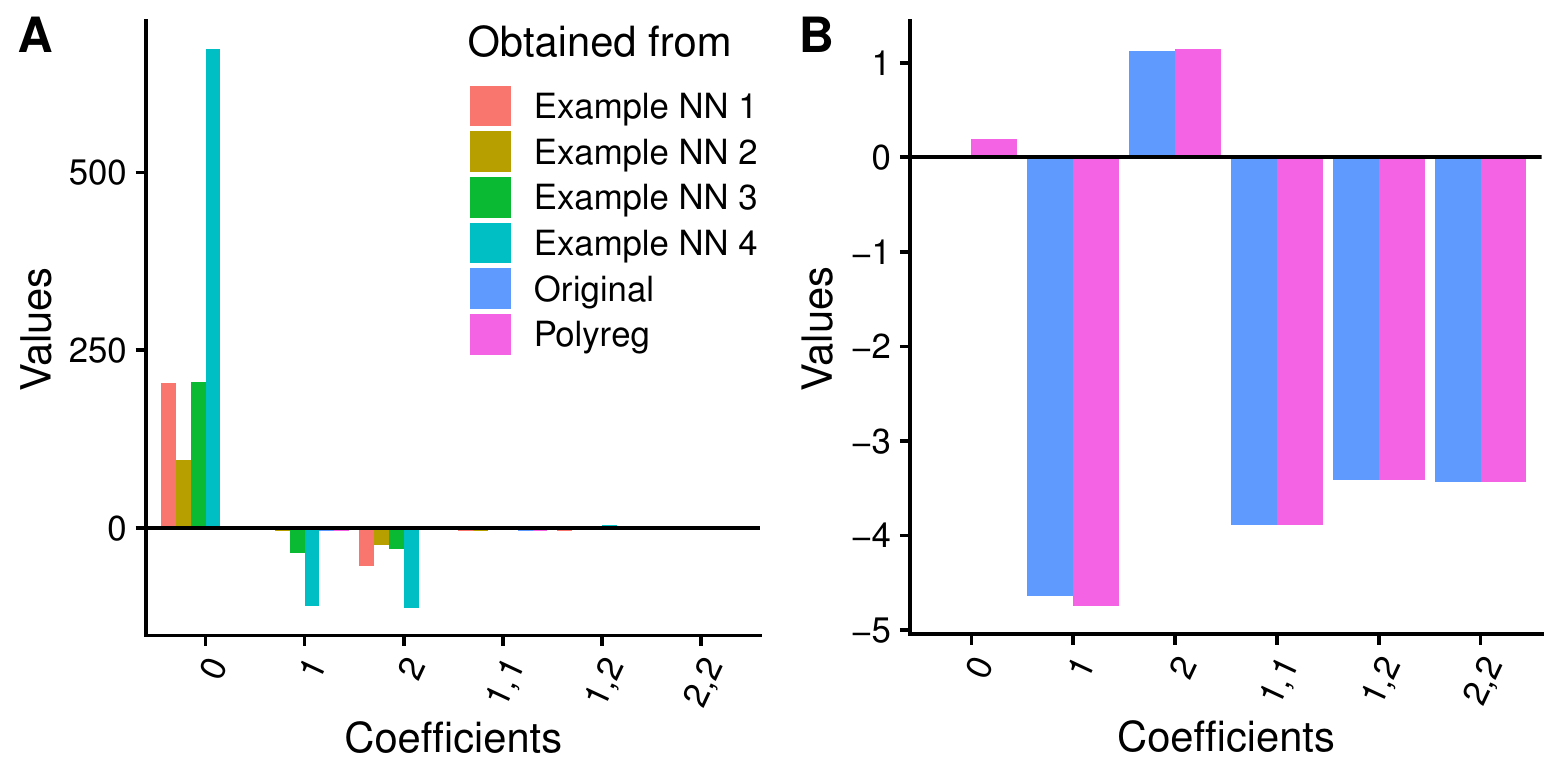}
	\caption{A: Comparison of the obtained coefficients with the proposed method applied to the four different NNs, the ones obtained by the \textit{polyreg} package and the original ones.}
	\label{coef_comparison}
\end{figure}

The four examples with different random seeds for the NN training initialization are represented in Fig.~\ref{coef_comparison_performance}. All of them have an acceptable accuracy, but, as it can be observed in the left side of Fig.~\ref{coef_comparison}, the obtained coefficients differ for each example. Furthermore, they have a huge difference with the original ones or the ones obtained by \textit{polyreg}. These two are represented on an appropriate scale in the right of Fig.~\ref{coef_comparison}, where it can be seen that, indeed, \textit{polyreg} obtained a much better approximation of the original coefficients.

\begin{figure}[ht]
	\centering
	\includegraphics[width = \textwidth]{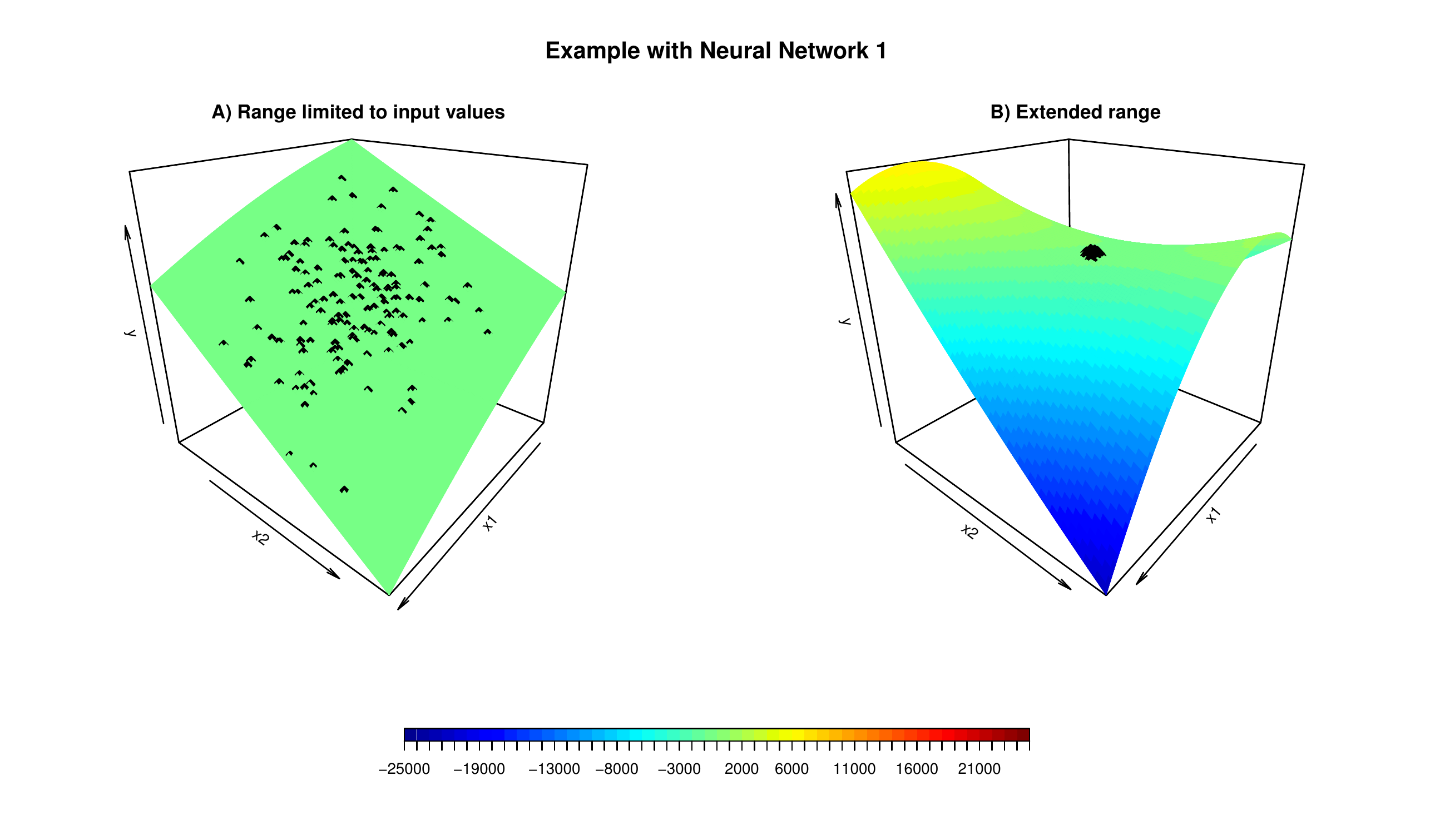}
	\caption{Polynomial surface obtained for NN example 1. A) Surface limited to input space. B) Surface extended over a larger space. The black dots represent the data points used in the training.}
	\label{surface_1}
\end{figure}

\begin{figure}[ht]
	\centering
	\includegraphics[width = \textwidth]{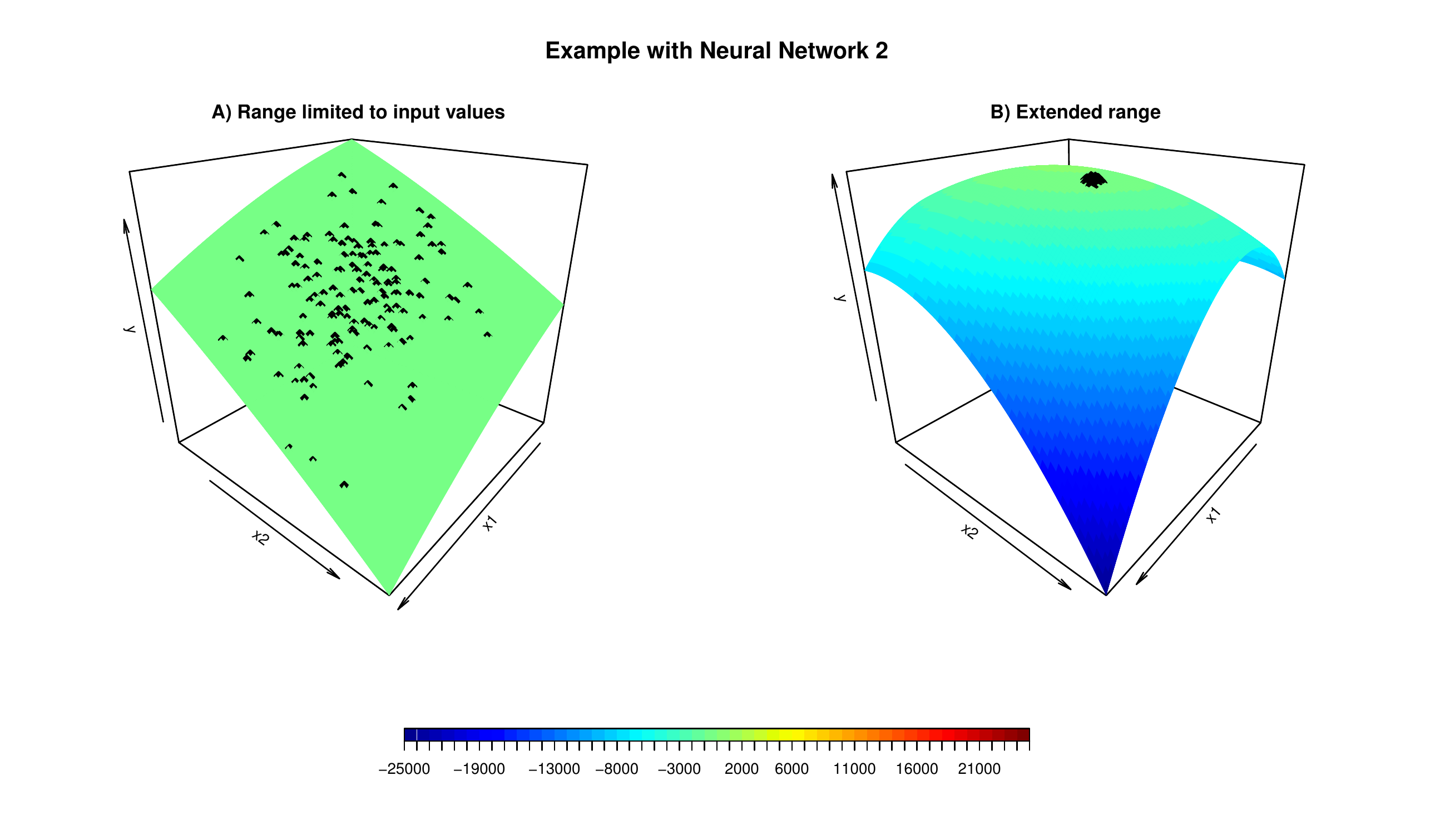}
	\caption{Polynomial surface obtained for NN example 2. A) Surface limited to input space. B) Surface extended over a larger space. The black dots represent the data points used in the training.}
	\label{surface_2}
\end{figure}

\begin{figure}[ht]
	\centering
	\includegraphics[width = \textwidth]{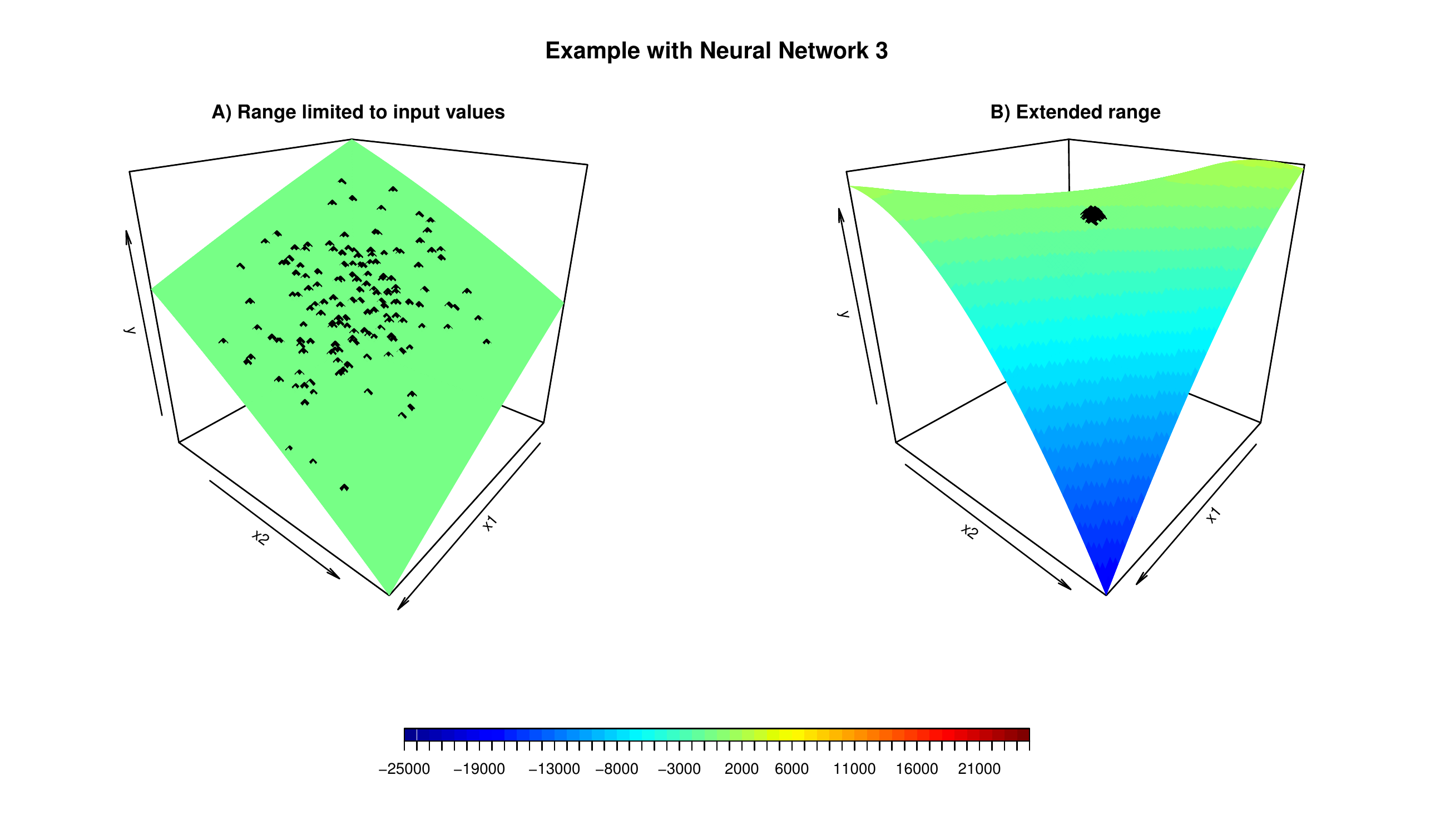}
	\caption{Polynomial surface obtained for NN example 3. A) Surface limited to input space. B) Surface extended over a larger space. The black dots represent the data points used in the training.}
	\label{surface_3}
\end{figure}

\begin{figure}[ht]
	\centering
	\includegraphics[width = \textwidth]{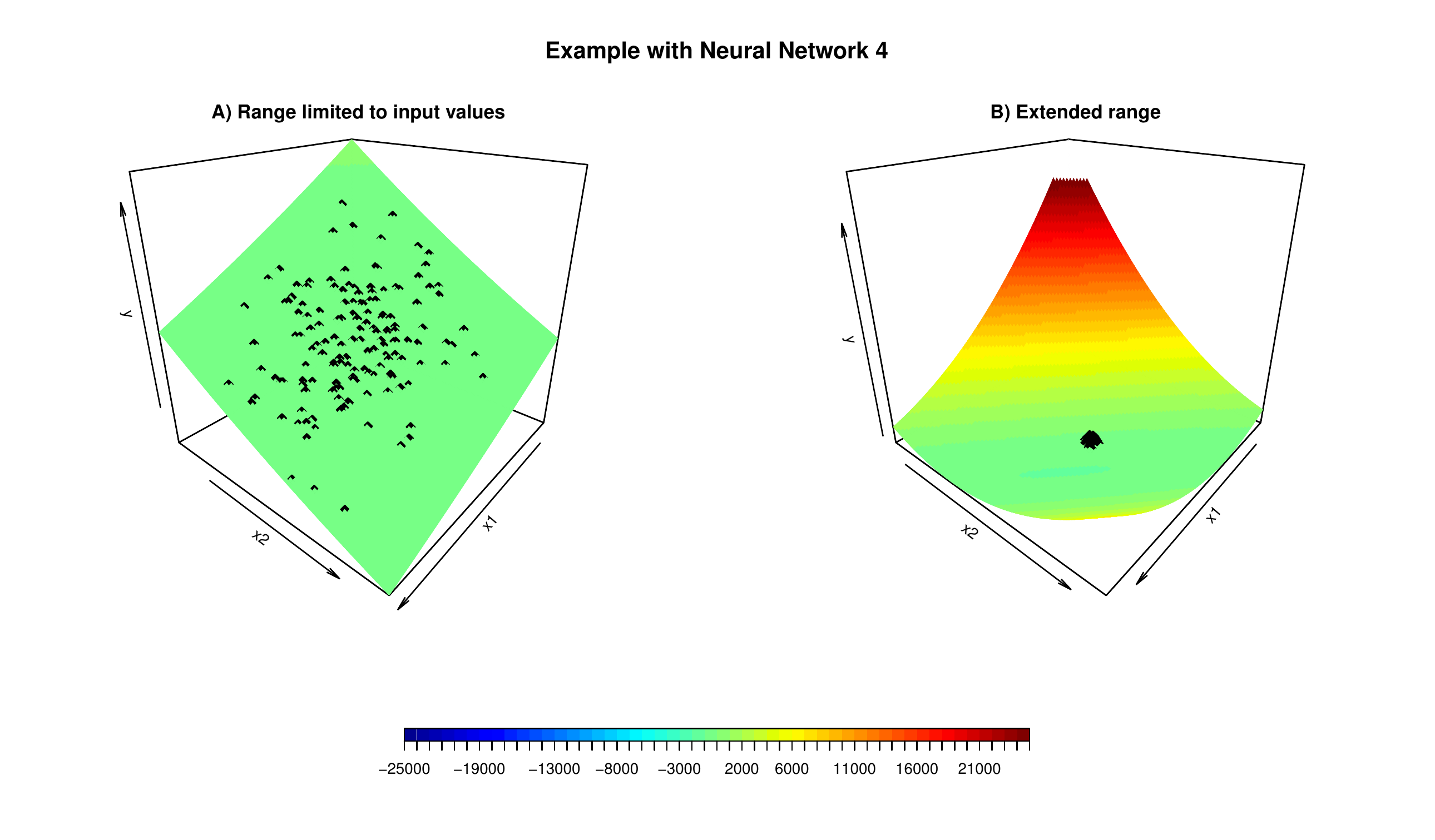}
	\caption{Polynomial surface obtained for NN example 4. A) Surface limited to input space. B) Surface extended over a larger space. The black dots represent the data points used in the training.}
	\label{surface_4}
\end{figure}

\begin{figure}[ht]
	\centering
	\includegraphics[width = \textwidth]{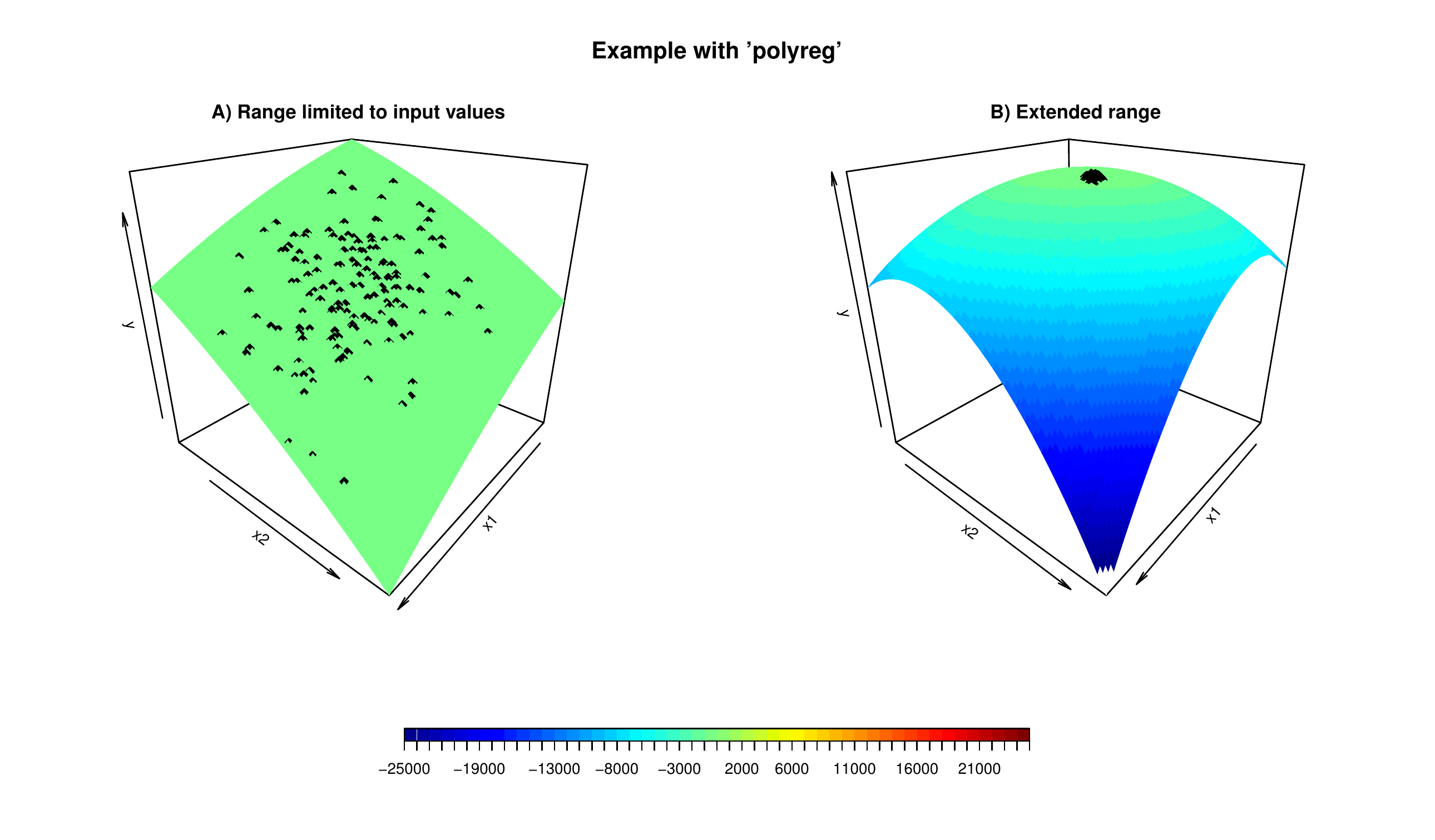}
	\caption{Polynomial surface obtained with the \textit{polyreg} package. A) Surface limited to input space. B) Surface extended over a larger space. The black dots represent the data points used in the training.}
	\label{surface_polyreg}
\end{figure}

\begin{figure}[ht]
	\centering
	\includegraphics[width = \textwidth]{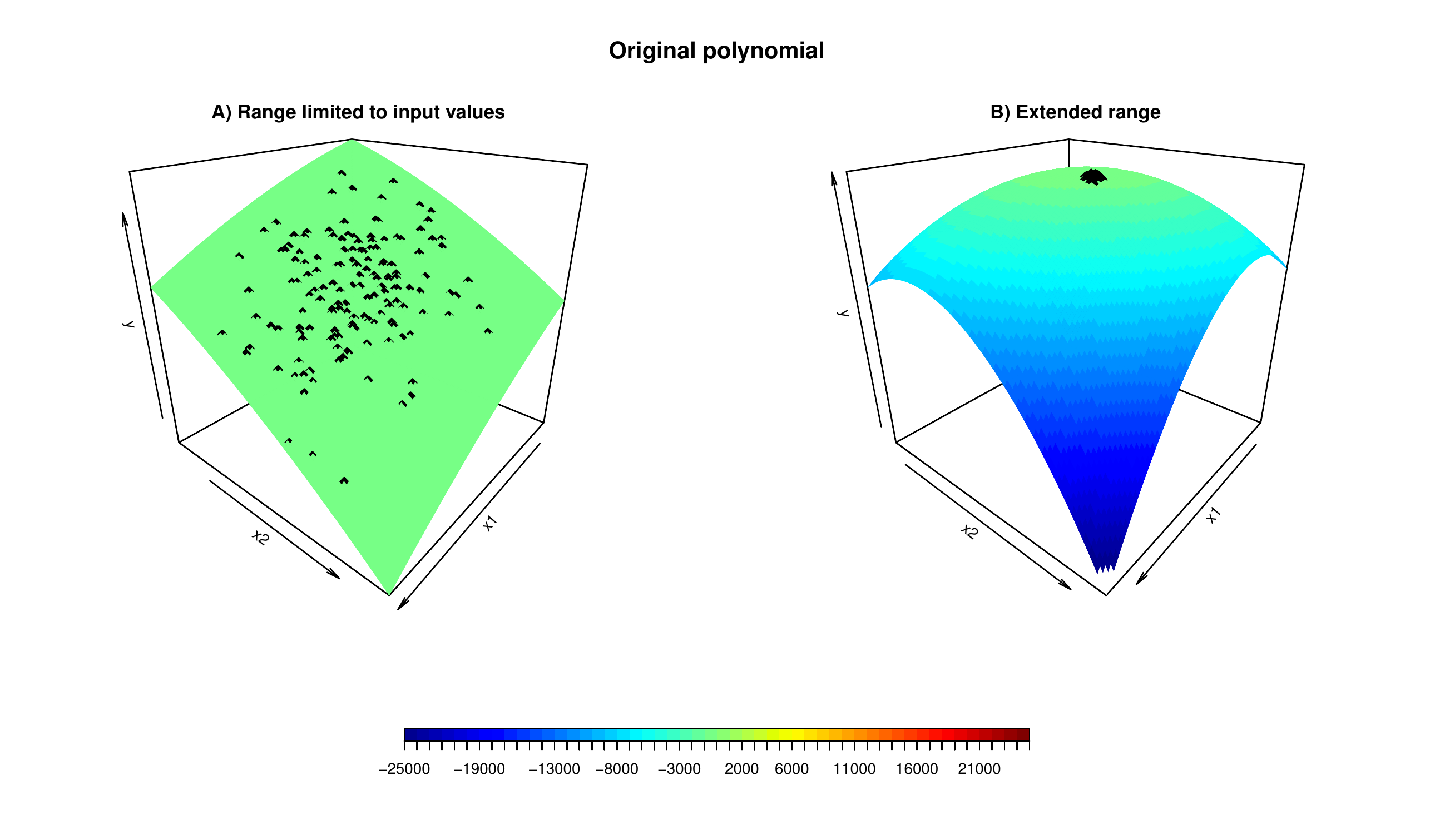}
	\caption{Polynomial surface obtained with the original coefficients. A) Surface limited to input space. B) Surface extended over a larger space. The black dots represent the data points used in the training.}
	\label{surface_original}
\end{figure}

The significant differences in the coefficients can be explained because, in the range in which the input data is contained, all of the obtained polynomials behave in a similar way. However, when the range is extended further away, the surfaces generated by the polynomials are completely different as it can be observed in Figures \ref{surface_1} to \ref{surface_4}. Clearly, the four NNs taken as example provide polynomials that in the extended range do not behave like the original one (Fig.~\ref{surface_original}), while the one obtained by \textit{polyreg} (Fig.~\ref{surface_polyreg}) is almost equal to the original one. This can be explained by the NN stopping after reaching a local minima in its optimization process. Furthermore, this shows that, in examples where the original data is actually polynomial, NNs can be precise in the input data region but fail when there is a need to extrapolate results outside of that region, while PR is able to learn a polynomial closer to the original one and can extrapolate results more accurately. This can be useful, for example, when learning physical laws that  could be polynomial, and where extrapolation can be useful. Finally, these findings are related with the idea of using simpler models in certain situations (\cite{rudinStopExplainingBlack2019}).

\section{Conclusions}\label{conclusion}

Several mathematical representations have been proposed in order to obtain a better understanding and interpretation of neural networks insights. Among them, polynomial regressions had been considered and explored to achieve this aim. However, under our knowledge, an explicit expression that could allow a deeper analysis had not been presented before.

The main contribution of this work is the explicit formula from Eq.~\eqref{betas_formula} to compute the coefficients of a polynomial regression that approximates the predictions of a single layer neural network, providing a mathematical framework to relate PR with NNs in a direct way, building one from the parameters of the other one. With respect to the study of the work made in \cite{chengPolynomialRegressionAlternative2019}, this shows that, at least for a single hidden layer, there is a clear relation between PR and NN like the authors conjectured and this can help transfer our knowledge of statistical PR into the field of NN. This framework in which we can explore the similarities of both models can help developing new ways of computing uncertainty estimations for NN or even the obtained coefficients could allow us to interpret the NN model using the classical statistical interpretation of the coefficients in regression models, helping to overcome the lack of interpretability of neural networks. Even more, through this framework, insights could be obtained about how to dimension and set the structure of NNs.

Even when a perfect fit is obtained when comparing the PR and NN predictions, this has some limitations. As it has been seen in the simulation experiments in Section \ref{experimental_results}, our formula depends highly on the input variables and the weights obtained in the NN, and when the final input value that the activation function receives is out of the acceptable range of the Taylor approximation, the obtained polynomial regression starts to fail. This could be solved by using some kind of regularization in the NN training that lets us have more control over the total value of the input of the activation function. Regarding this problem, it is also found that, for the simple examples simulated, lower number of neurons in the hidden layer produce slightly better results than higher ones, meaning that for more complex NNs this method should be improved. 

Concerning the extension of this formula to more general problems, extending it to a NN with more output units is trivial as it is just required to use the same formula for each of the desired output units. Furthermore, if a classification problem is involved, a non linear activation function must be considered to have on each output unit $z_k=\hat{g}\left(\sum_{j=0}^{h_1}v_{kj}y_j\right)$  where $k$ denotes the output unit. Then, Taylor's theorem should be used again and the problem to solve is similar to what should be solved when extending the method to NNs with two hidden layers. It is important to note here that this extension should be made carefully, because the approximation of a deeper layer would depend on how well the previous layer is approximated, on the number of neurons of the previous layers and on the number of input variables. 

In terms of the Taylor approximation, it is interesting to see that increasing the order does not strictly increase  its performance, as it increases the asymptotic behavior of the error in the extremes. However, it is remarkable that when using an approximation of higher order than the polynomial that originated the data, the coefficients of terms that exceed the original order tend to be shrunk to zero. Also, the examples presented use low dimensional data, because with higher dimensions the method starts to fail, being out of the approximation range of the activation function. Therefore, improvements could be made to the Taylor expansion approach in an attempt to increase the range of values in which the approximation is valid. In this context, extending this method for piecewise activation functions could be explored by means of piecewise polynomial approximations of those functions.

\section*{Acknowledgments}
This research has been partially supported by Ministerio de Economía, Industria y Competitividad, Gobierno de España, grant number PID2019-104901RB-I00 and PID2019-106811GB-C32.

\bibliography{mybibfile}

\end{document}